\documentclass[conference]{IEEEtran}
\IEEEoverridecommandlockouts
\usepackage{cite}
\usepackage{amsmath,amssymb,amsfonts}
\usepackage{algorithmic}
\usepackage{graphicx}
\usepackage{textcomp}
\usepackage{xcolor}
\def\BibTeX{{\rm B\kern-.05em{\sc i\kern-.025em b}\kern-.08em
    T\kern-.1667em\lower.7ex\hbox{E}\kern-.125emX}}
\begin{document}

\title{Frequency-aware Neural Representation for Videos}

\author{Jun Zhu \qquad
    Xinfeng Zhang\qquad
    Lv Tang \qquad
    Junhao Jiang \qquad
    Gai Zhang\qquad
    Jia Wang\\
    University of Chinese Academy of Sciences\\
    {\tt\small \{zhujun23, xfzhang, jiangjunhao25, zhanggai16, wangjia242\}@mails.ucas.ac.cn,}\\
    {\tt\small luckybird1994@gmail.com}\\}
\maketitle

\begin{abstract}
Implicit Neural Representations (INRs) have emerged as a promising paradigm for video compression. However, existing INR-based frameworks typically suffer from inherent spectral bias, which favors low-frequency components and leads to over-smoothed reconstructions and suboptimal rate–distortion performance.
In this paper, we propose FaNeRV, a Frequency-aware Neural Representation for videos, which explicitly decouples low- and high-frequency components to enable efficient and faithful video reconstruction. FaNeRV introduces a multi-resolution supervision strategy that guides the network to progressively capture global structures and fine-grained textures through staged supervision . To further enhance high-frequency reconstruction, we propose a dynamic high-frequency injection mechanism that adaptively emphasizes challenging regions. In addition, we design a frequency-decomposed network module to improve feature modeling across different spectral bands.
Extensive experiments on standard benchmarks demonstrate that FaNeRV significantly outperforms state-of-the-art INR methods and achieves competitive rate-distortion performance against traditional codecs. 
\end{abstract}


\section{Introduction}

Video content has become the dominant contributor to Internet traffic, placing tremendous pressure on storage and transmission resources. Video compression alleviates this burden by significantly reducing data volume while preserving satisfactory visual fidelity.
Conventional hybrid video coding methods have achieved remarkable rate-distortion performance by exploiting both spatial and temporal redundancies.
The latest hybrid video coding standard, Versatile Video Coding (H.266/VVC), delivered a great improvement in compression efficiency. It achieves comparable visual quality at roughly half the bit rate of the widely adopted standard H.265/HEVC.
Despite these achievements, the rigid structure of sequentially connected handcrafted modules, such as block partitioning, linear transformation, and quantization, limits hybrid codecs from achieving global optimality. 
Since each component is typically optimized in isolation using heuristic rules, end-to-end rate–distortion optimization remains challenging, motivating the exploration of neural video compression methods.

Implicit Neural Representations (INRs) have recently emerged as a powerful paradigm for data compression.
By formulating data representation as a continuous function approximation problem, INRs map input coordinates directly to signal attributes via a neural network.
This approach enables holistic modeling through global optimization over the entire data instance.
Specifically, in the context of video compression, a sequence is treated as a continuous function that maps spatiotemporal coordinates to pixel values.
Compression is achieved by training the network to overfit the target video while minimizing reconstruction distortion.
The learned parameters are then quantized and entropy-coded, producing a compact bitstream.
Several frameworks have been proposed to leverage INRs for video compression, achieving significant improvements in coding efficiency.
Early works, such as NeRV (Neural Representations for Video ~\cite{chen2021nerv}), treat video as a function of frame indices, successfully achieving compression ratios comparable to widely used standards like H.264/AVC and H.265/HEVC. 
Subsequent advancements, including H-NeRV~\cite{chen2023hnerv} and HiNeRV~\cite{kwan2024hinerv}, further enhance rate–distortion performance through architectural innovations such as content-adaptive embeddings and hierarchical feature grids.
\begin{figure}[tp]
    \centering
    \includegraphics[width=\linewidth]{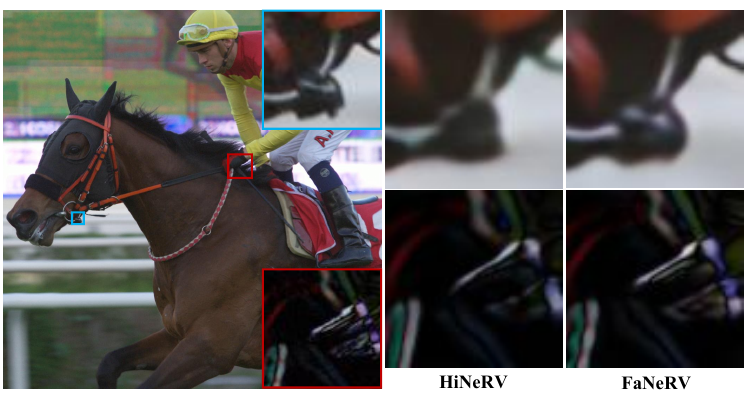}
    \caption{ Reconstruction results on video detail regions and motion areas. The motion area (red box) is visualized by computing the difference between the current and subsequent frame. Reconstruction details are shown in the region marked by the blue box.}
    \label{fig:dynamic}
\end{figure}

Despite their success, INRs tend to converge slowly on high-frequency details due to the inherent spectral bias of neural networks. In texture-rich scenes, this often results in over-smoothing, where fine-grained details, such as hair, fabric textures, or water surfaces, are attenuated.
The limitation stems from the spectral complexity of natural videos, which comprise a mixture of diverse frequency modes.
When an INR attempts to model the entire spectrum simultaneously, the entanglement of low and high frequency signals leads to an ill-conditioned optimization landscape. Dominant low-frequency components can overshadow the gradients required for high-frequency convergence, destabilizing the training process and, in extreme cases, causing training collapse.
This observation suggests that video representation should be decomposed across multiple frequency scales rather than modeled holistically.

To overcome the spectral coupling limitation, we propose a Frequency-aware Neural Representation for Video, FaNeRV.
The framework incorporates multi-resolution supervision to reformulate video reconstruction as a staged, progressive cascaded process.
At the shallow stages,
the network is supervised using downsampled low-resolution signals,
which act as low-pass filtered versions of the original video and encourage the network to focus on capturing global structures.
As the signal propagates to deeper layers, the supervision resolution is gradually increased, enabling subsequent stages to concentrate on modeling high-frequency residuals and recovering fine-grained details.
This frequency-aware design explicitly decouples low and high frequency components, ensuring faithful and efficient reconstruction of the full spectral content.

Furthermore, to enhance the recovery of fine-grained details, we introduce a dynamic high-frequency injection mechanism.
Unlike static edge supervision, which uniformly targets all high-frequency components, our approach selectively emphasizes the most challenging regions by extracting high-frequency signals from the real-time reconstruction residuals.
This mechanism enables the network to progressively shift its focus from coarse boundaries in early stages to subtle micro textures as training progresses. Such adaptive guidance ensures that the network’s capacity is concentrated on the most critical missing details.
In addition, we introduce a network module that effectively decomposes features across different frequency bands within the network.

To validate the superiority of our proposed framework, extensive experiments are conducted on standard benchmarks.
Quantitative results demonstrate that FaNeRV significantly outperforms existing INR-based and learning-based video compression methods in terms of Rate-Distortion (RD) performance.
Notably, FaNeRV even surpasses VTM, which is widely recognized as the most efficient traditional coding standard.
Qualitative comparisons in Fig~\ref{fig:dynamic} further show that our frequency-aware decoupling strategy not only minimizes statistical distortion but also produces visually coherent reconstructions that align better with human perceptual preference.

In summary, our main contributions are as follows:
\begin{itemize}
    \item We propose FaNeRV, a frequency-aware implicit compression framework that integrates Multi-Resolution Supervision and Dynamic High-Frequency Injection.
    \item Based on FaNeRV, we implement an INR video codec incorporating a block that adaptively decomposes low- and high-frequency features, enhancing the network’s capacity for fine-grained reconstruction.
    \item Extensive experiments validate the effectiveness of our method, demonstrating its potential to surpass existing state-of-the-art approaches. Furthermore, the framework provides robust support for scalable video coding.
\end{itemize}
\begin{figure*}[ht]
\centering
\includegraphics[width=\linewidth]{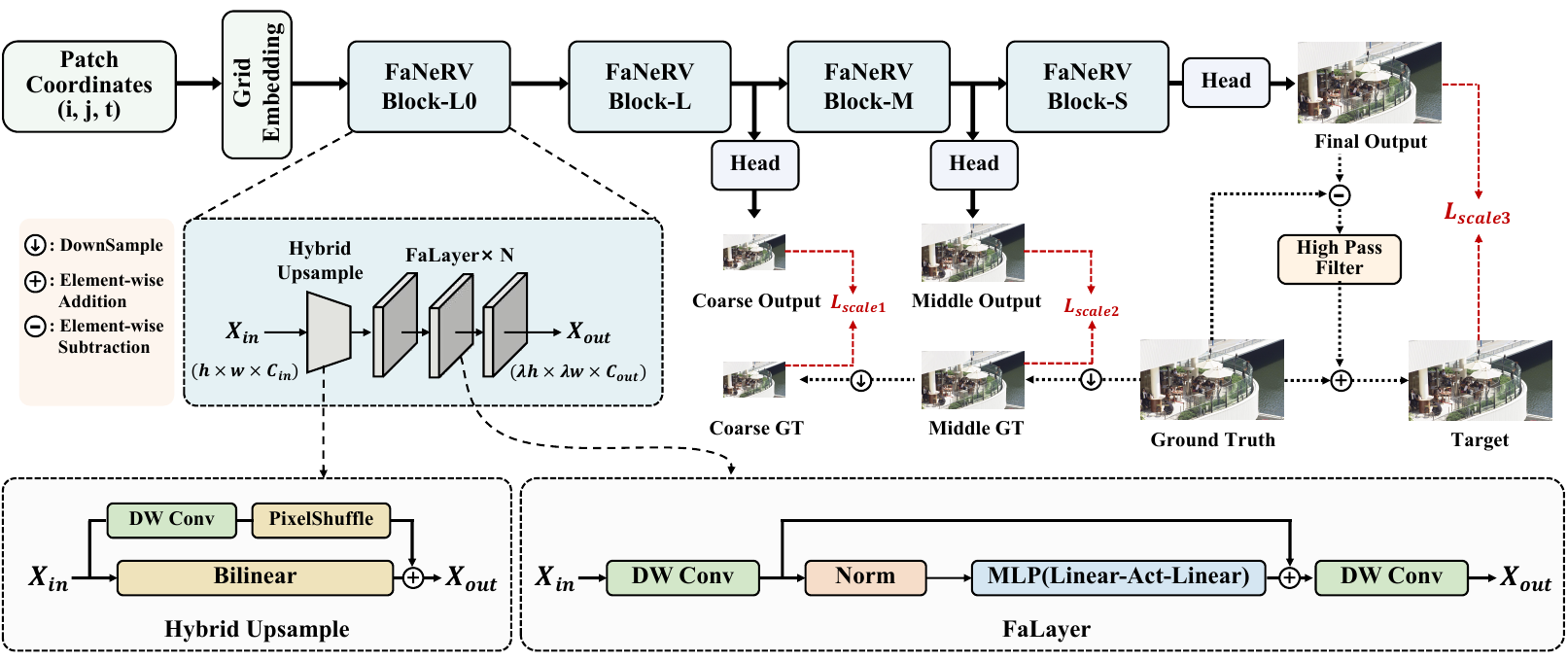} 
\caption{Overview of the proposed FaNeRV architecture. A coordinate input retrieves a grid embedding. This embedding passes through blocks with reduced parameters, upsampling in resolution. Outputs at different depths are supervised by the original image and its downsampled versions.
Details of these blocks are shown in the bottom.}
\label{fig:main}
\end{figure*}
\section{Related Work}
\subsection{Implicit Neural Network}
Implicit neural representation (INR) is a novel approach in the field of data representation. Continuous signals, such as images \cite{dupont2021coin}, videos \cite{Tang_2023_ICCV, chen2023hnerv,kwan2024hinerv} and 3D scenes \cite{mildenhall2021nerf}, are encoded using neural networks.
INRs parameterize a signal as a function that maps coordinates to corresponding values.
Neural representation for videos (NeRV) \cite{chen2021nerv} encodes videos by training a neural network to output the entire frame when given a frame index. This approach transforms video compression into a model compression task, where the video content is encapsulated within the network's parameters.
The following contributions further explore content-adaptive structure \cite{chen2023hnerv}, conditional decoder \cite{zhang2024boosting} and hierarchical positional encoder \cite{kwan2024hinerv} for efficient video compression.

\subsection{Spectral Bias}
Extensive research has demonstrated that deep neural networks exhibit a spectral bias, meaning they tend to learn low-frequency components of a target function much faster than high-frequency ones~\cite{rahaman2019spectral}.
In the context of Implicit Neural Representations (INRs), this bias becomes a significant bottleneck: networks prioritize global shapes and smooth color transitions, but struggle to capture fine-grained details.
To mitigate this, pioneering works like NeRF \cite{mildenhall2021nerf} introduce Positional Encoding (PE), which maps low-dimensional input coordinates into a higher-dimensional frequency space using high-frequency sinusoidal functions.
Similarly, SIREN \cite{sitzmann2020implicit} proposed periodic sine activation functions to propagate gradients for high-frequency signals.  
Subsequent studies have explored high-frequency loss functions, such as Focal Frequency Loss \cite{jiang2021focal}, which explicitly penalize errors in the frequency domain to force the network to attend to missing details.

Despite these advancements, most existing strategies rely on static frequency mappings or fixed loss weights. They often fail to consider the dynamic nature of video signals, where the frequency distribution can vary significantly across frames and network depths. 
In contrast, our work addresses spectral bias using a dynamic strategy, enabling the network to adaptively focus on missing high-frequency components.

\section{Methods}
\subsection{Network Architecture}
FaNeRV adopts a patch-wise learning paradigm to balance training efficiency and reconstruction quality. The implicit network takes the spatiotemporal coordinate of a patch as input and outputs the corresponding pixel tensor for the entire region. Formally, the video is modeled as a function $f: (x_p, y_p, t) \to \mathbb{R}^{H_p\times W_p \times 3}$.
As illustrated in Fig.~\ref{fig:main}, input coordinates are first mapped into high-dimensional embeddings via multi-resolution learnable feature grids, following the design in FFNeRV~\cite{lee2023ffnerv}.
These embeddings are then processed by a cascade of encoding blocks that integrate a hybrid upsampling module with Frequency-aware layers (Falayer) to refine feature representations.
Finally, multiple lightweight projection heads are attached at intermediate and final stages of the network to map the refined features back to the RGB space, producing video outputs at multiple resolutions.
\subsubsection{Hybrid Upsample}
In the feature processing stage, we introduce a hybrid upsampling module to recover the spatial resolution. This module consists of two parallel pathways: a non-parametric branch utilizing bilinear interpolation, and a learnable branch employing PixelShuffle combined with a depthwise convolution.
Given an input feature $X_{in}$, the two branches process the signal independently, and their outputs are fused via element-wise addition to generate the up-scaled feature $X_{out}$.
The bilinear interpolation efficiently reconstructs the global structure without introducing additional parameters, serving as the low-frequency prior. This allows the PixelShuffle branch to focus on capturing the high-frequency residuals.
Consequently, the learnable branch can ultilize lightweight convolutions, effectively minimizing model complexity while ensuring high-quality reconstruction.
\subsubsection{Falayer}
Existing INR blocks typically utilize stacked 3$\times$3 convolutions or MLPs. However, such local operations limit the receptive field, hindering the capture of long-range spatial dependencies. Although increasing the network depth can theoretically enlarge the receptive field, it often leads to excessive computational costs. 
To address this limitation, we propose a Frequency-aware layer based on the ConvNeXt block.
This block combines large-kernel depthwise convolutions for broad spatial perception with MLP layers for semantic feature transformation, and is normalized using Layer Normalization.
Furthermore, we introduce a residual connection from the output of the depthwise convolution to the MLP output. This design preserves the spatial information captured by the large kernel and improves training stability.
\subsection{Multi-Resolution Supervision}
During the training phase, standard INR-based video compression frameworks typically optimize the network by minimizing a single reconstruction loss between the final output and the ground-truth video.
However, due to the inherent spectral bias of deep neural networks, low-frequency components are learned much faster than high-frequency details. 
As a result, jointly optimizing both global structures and fine textures with a single end-to-end loss becomes inefficient. The dominant low-frequency components tend to overshadow the learning of high-frequency details. This imbalance leads to over-smoothed reconstructions and constrains the rate-distortion performance of the model.

We introduce a Multi-Resolution Supervision to disentangle the modeling of different frequency components.
Specifically, lightweight projection heads are attached to intermediate stages of the network. Let \{$X_1, X_2,\dots,X_S$\} denote the feature maps at varying depths, where the spatial resolution increases progressively. At each stage $i$, a projection head maps $X_i$ to an intermediate video prediction $\hat{V_i}$.
Simultaneously, we generate a pyramid of ground truth videos \{$V_1, V_2, \dots,V_{S-1}$\} by cascaded downsampling the original video $V_S$.
We employ max pooling for downsampling because it is invariant to small translations, improving the stability of the patch-wise training.

\begin{table*}[t]
\centering

\caption{
Video regression results (PSNR$\uparrow$) and model complexity on HEVC ClassB \cite{sullivan2012overview} dataset.
}
\begin{tabular}{c c c c c c c c c c c}

\noalign{\hrule height 1.2pt}
\noalign{\vskip 1pt}
 Model&Size& MACs &Enc. FPS &Dec. FPS&Bas. & BQT. & Cac. & Kim. & Par. &Avg.\\
\hline
\noalign{\vskip 1pt}
HNeRV-Boost&3.05M& 206.3G& 34.6 img/s&71.9 img/s&28.49&27.79&30.09&32.55&29.71&29.63\\
HiNeRV&3.15M & 187.2G& 26.5 img/s&58.1 img/s&31.43&31.19&31.46&34.41&31.25&31.95\\
FaNeRV&3.03M & 183.1G& 20.3 img/s&46.7 img/s&\textbf{32.68}&\textbf{32.27}&\textbf{32.70}&\textbf{35.80}&\textbf{32.75}&\textbf{33.24}\\
\hline
\noalign{\vskip 1pt}
HNeRV-Boost&6.62M& 360.9G& 25.4 img/s&54.3 img/s&31.17&30.31&32.71&35.09&32.04&32.26\\
HiNeRV&6.42M&368.2G& 21.9 img/s&45.9 img/s&33.45&32.54&33.14&36.41&32.86&33.68\\
FaNeRV&5.92M& 348.0G&16.8 img/s&37.9 img/s&\textbf{34.30}&\textbf{33.16}&\textbf{34.09}&\textbf{37.34}&\textbf{34.50}&\textbf{34.68}\\
\noalign{\hrule height 1.2pt}
\end{tabular}

\label{tab:regression}
\end{table*}

This supervision strategy provides progressive structural guidance. By imposing constraints at lower resolutions, shallow layers are guided to capture low-frequency structures, including overall shapes and spatial layouts.
As the network deepens and supervision is applied at higher resolutions, deeper layers can refine high-frequency textures while leveraging the stable structural context established by earlier layers.
Since low-frequency components carry most of the spectral energy and serve as a fundamental prior for high-frequency reconstruction, a larger fraction of model parameters is allocated to the shallow layers to ensure sufficient capacity for low-frequency modeling.
This multi-resolution supervision mechanism stabilizes training by constraining the optimization space with intermediate guidance and improves the overall reconstruction quality.

\subsection{Dynamic High-Frequency Injection}

Despite the structural decoupling achieved by Multi-Resolution Supervision, reconstructing fine details remains challenging, as the optimization at deep layers still favor low-frequency smoothness. To mitigate this, we introduce a Dynamic High-Frequency Injection mechanism, which dynamically incorporates high-frequency information into the training objectives, enhancing the reconstruction of sharp details.

Unlike prior approaches that employ either static frequency supervision or additional high-frequency modules, our method leverages a dynamic high-frequency injection strategy.
Specifically, we compute the residual between the generated video $\hat{V}_S$ and the original video $V_{S}$. A high-pass filter $\mathcal{H}$ is then applied to extract high-frequency components from this residual.
\begin{equation}
    R_{HF}=\mathcal{H}(V_{S}-\hat{V}_S).
\end{equation}
This term $R_{HF}$ represents the fine details that the network currently fails to capture. We inject this high-frequency residual back into the original ground truth to construct an enhanced target $V_{target}$:
\begin{equation}
    V_{target}=V_{S}+\beta \cdot R_{HF},
\end{equation}
where $\beta$ is a scalar controlling the injection intensity. By optimizing against this enhanced target, we increase the penalty for missing high-frequency details, forcing the network to prioritize the learning of these challenging textural patterns.

\subsection{Loss Function}
To ensure high-fidelity reconstruction across all resolutions, we adopt a scale-adaptive loss (SA Loss) as the loss function for FaNeRV. It combines the L1, MSE and MS-SSIM losses:
\begin{equation}
    \mathcal{L}_{SA}^{(r)} = \alpha_r \mathcal{L}_{MSE}^{(r)}+\beta_r \mathcal{L}_{1}^{(r)} +(1-\alpha_r -\beta_r)\mathcal{L}_{MS-SSIM}^{(r)},
\end{equation}
where $\alpha_r$ and $\beta_r$ are coefficients determined by the resolution level $r$.
Notably, since the MS-SSIM metric primarily focuses on the perceptual similarity that is crucial for the final reconstructed video, it is exclusively applied at original resolution. 

Given the formulas of $\mathcal{L}_{MSE}$ and $\mathcal{L}_1$: 
\begin{equation}
    \mathcal{L}_{MSE}^{(r)}=\frac{1}{T}\sum_{t=1}^T||\hat{V}^{t}_r-V^t_r||_2^2, 
    \mathcal{L}_{1}^{(r)}=\frac{1}{T}\sum_{t=1}^T||\hat{V}^{t}_r-V^{t}_r||_1,
\end{equation}
we observe that $\mathcal{L}_{MSE}$ is more sensitive to large errors while $\mathcal{L}_1$ is more robust with noisy data.
Prior studies~\cite{zhao2016loss} also indicate that $\mathcal{L}_{MSE}$ is effective for global structural stability and tends to produce over-smoothed results, whereas $\mathcal{L}_1$ better retains sharp edges and high-frequency details.
Based on discussions above, FaNeRV increases $\alpha_r$ at lower resolutions to establish a stable skeleton, while increasing $\beta_r$ at high resolution to ensuring fine textures.

The total loss in FaNeRV is the sum of losses across all scales, with the high-frequency enhanced video serving as the reference at the original resolution:
\begin{equation}
    \mathcal{L}_{total}=\sum_{r=1}^{S-1}\mathcal{L}_{SA}^{(r)}(\hat{V}_{r},V_r)+\mathcal{L}_{SA}^{(S)}{(\hat{V}_{S},V_{target})}.
\end{equation}

\section{Experiments}
\label{sec:experiments}

\subsection{Video Representation}

\begin{table*}[t]
\centering
\caption{
BD-Rate results relative to FaNeRV on HEVC ClassB \cite{sullivan2012overview} and UVG \cite{mercat2020uvg} datasets.
}
\begin{tabular}{c c c c c c c c}
\noalign{\hrule height 1.2pt}
\noalign{\vskip 1pt}
Dataset&Metric&HiNeRV&HNeRV-Boost&DCVC-DC&DCVC-FM&VTM(RA)&HM\\
\hline
\noalign{\vskip 1pt}
ClassB&PSNR&$-42.2\%$& $-74.1\%$& $-23.2\%$&$-27.0\%$&$-6.1\%$&$-35.4\%$\\
                    &MS-SSIM&$-27.5\%$&$-86.4\%$ & $-43.8\%$&$-45.5\%$&$-27.1\%$&$-59.6\%$\\
\hline
\noalign{\vskip 1pt}
UVG&PSNR&$-26.9\%$&$-65.4\%$ &$-6.3\%$&$-10.8\%$&$7.7\%$&$-32.1\%$\\
                    &MS-SSIM&$-16.3\%$&$-75.1\%$ &$-29.8\%$&$-32.9\%$&$-25.5\%$&$-53.3\%$\\
\noalign{\hrule height 1.2pt}
\end{tabular}
\label{tab:compression}
\end{table*}
\begin{figure*}[t]
\centering
\includegraphics[width=\linewidth]{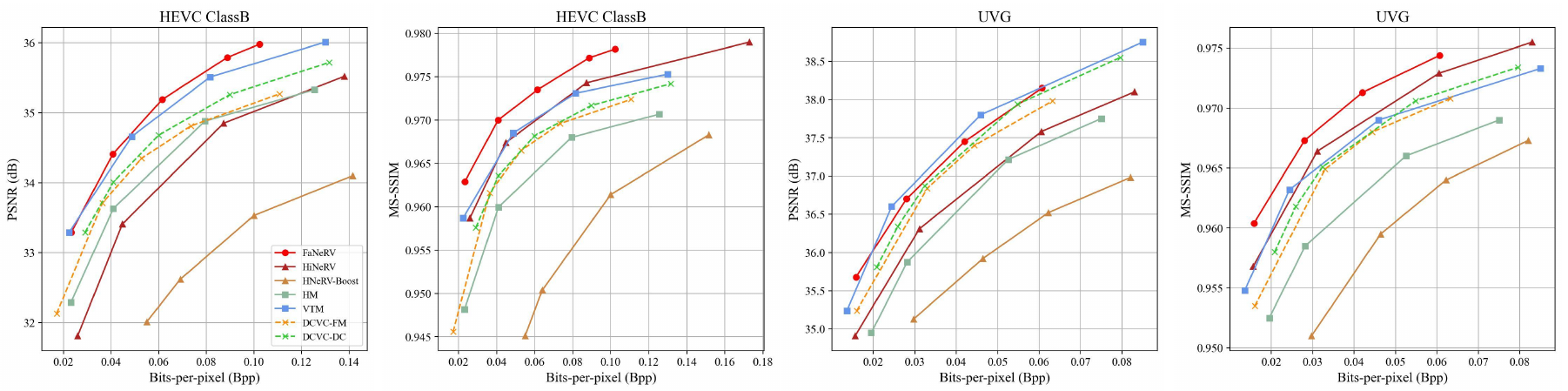} 

\caption{Video compression results on HEVC ClassB \cite{sullivan2012overview} and UVG \cite{mercat2020uvg}.}
\label{fig:compression}
\end{figure*}


\begin{table}[t]
\centering

\caption{
Complexity comparison on the BQTerrace. The INR-based models and DCVC series are
evaluated on NVIDIA 3090 GPUs.
}
\begin{tabular}{c | c c c c}
\noalign{\hrule height 1.2pt}
 Method&Encoding time($\downarrow$) &  Decoding FPS ($\uparrow$)\\
 \hline
HM& 2h49min &16.7 img/s\\
VTM &9h48min &11.3 img/s\\
DCVC-DC  &6min & 2.1 img/s\\
DCVC-FM &6min & 1.9 img/s\\
HNeRV-boost &1h30min& 63.9 img/s\\
HiNeRV     &2h13min& 54.3 img/s\\
FaNeRV &2h54min &46.2 img/s\\
\noalign{\hrule height 1.2pt}
\end{tabular}

\label{time}
\end{table}

To evaluate the representation capability of FaNeRV, we conduct experiments on five video sequences with a resolution of 1920$\times$1080 from HEVC ClassB \cite{sullivan2012overview} dataset,
all of which are captured from real-world scenes.
We compare FaNeRV with representative INR-based methods, including HiNeRV \cite{kwan2024hinerv} and HNeRV-boost \cite{zhang2024boosting}.
For a fair comparison, all models are configured with comparable parameter sizes. We use open-source implementations of these methods and adhere to their default settings.
All models are trained for 300 epochs, and the Peak Signal-to-Noise Ratio (PSNR) metric is used to evaluate reconstruction quality.

Tab. \ref{tab:regression} summarizes the model sizes and representation performance of all compared methods across test sequences.
Constrained by the inherent spectral bias of deep networks, existing INR-based methods suffer from severe motion blur and over-smoothing, particularly in sequences with high temporal dynamics, such as \textit{BasketballDrive} and \textit{BQTerrace}. 
However, by incorporating Multi-Resolution Supervision and Dynamic High-Frequency Injection, FaNeRV allocates sufficient representational capacity to temporal complex regions, effectively mitigating the spectral inertia that limits competing methods.
Moreover, benefiting from the low-frequency anchoring provided by the large-kernel backbone, FaNeRV exhibits superior geometric consistency and color fidelity in relatively static scenes like \textit{ParkScene} and \textit{Kimono1}.
Visualization results in Fig. \ref{fig:dynamic} further confirm that FaNeRV not only recovers richer high-frequency details than HiNeRV but also achieved improved temporal stability.

\subsection{Video Compression}

We compare FaNeRV with traditional codecs (e.g., H.266/VTM\cite{browne2022algorithm} , H.265/HM), learning-based approaches (e.g., DCVC-DC \cite{li2023neural}, DCVC-FM \cite{li2024neural}), as well as INR-based methods (e.g., HiNeRV, HNeRV-boost) on HEVC ClassB \cite{sullivan2012overview} and UVG \cite{mercat2020uvg} datasets.
The performance of DCVC-DC and DCVC-FM is evaluated using the official implementation and pre-trained models provided by the authors.
INR-based methods employ their respective strategy to compress network parameters.
For our method, we further perform 30 epochs of Quantization-Aware Training (QAT) \cite{jacob2018quantization} to reduce quality loss introduced by quantization. Entropy coding \cite{mentzer2019practical} is then applied to generate the compressed bitstream.
Reconstruction quality is evaluated using PSNR and MS-SSIM, where MS-SSIM is used to assess perceptual quality. Compression efficiency is measured in terms of bits per pixel (bpp), and Bj\o ntegaard Delta bitrate \cite{bjontegaard2008improvements} (BD-Rate) is employed to quantify bitrate savings across different codecs.

Table \ref{tab:compression} reports the BD-rate results, reflecting the bitrate savings or increases of our method relative to competing approaches at equivalent video quality. FaNeRV outperforms the state-of-the-art conventional codec, VTM (RA) on the HEVC ClassB \cite{sullivan2012overview} dataset, which contains more dynamic video content. This result highlights the suitability of FaNeRV for real-world scenes with complex motion.
On the UVG dataset, FaNeRV achieves performance close to that of VTM (RA), demonstrating its robustness across a diverse range of video content.
Compared to HiNeRV, which employs a similar network parameter coding strategy, FaNeRV achieves a bitrate saving of $43.4\%$ on the HEVC ClassB dataset and a $28.8\%$ reduction on the UVG dataset in terms of PSNR. These results indicate that FaNeRV provides superior representation efficiency among INR-based video coding methods.

Fig. \ref{fig:compression} illustrates rate-distortion curves on UVG\cite{mercat2020uvg} and HEVC ClassB\cite{sullivan2012overview} datasets.
FaNeRV consistently achieves strong performance across a wide range of bitrates.
In addition, FaNeRV outperforms all competing approaches under the MS-SSIM metric, indicating improved perceptual quality of reconstructed videos.

Tab.~\ref{time} presents the encoding and decoding time of various coding schemes. FaNeRV exhibits a modest increase in runtime compared to other INR-based approaches, which is a reasonable trade-off given the notable gains in rate–distortion performance. Furthermore, FaNeRV preserves the fast decoding property inherent to INR-based methods and achieves significantly lower decoding latency than both conventional codecs and learning-based approaches.

\subsection{Scalable coding}

\begin{table}[t]
\centering
\caption{
Scalable coding results on HEVC ClassB \cite{sullivan2012overview}. HiNeRV adopts separate models with sizes of 3.1M, 3.1M, and 3.3M for different resolutions. FaNeRV trains a single model of 3.3M.
}
\begin{tabular}{c | c  c }
\noalign{\hrule height 1.2pt}
& HiNeRV & FaNeRV\\
\hline
480x270& 30.84dB@0.278bpp& 31.94dB@0.271bpp\\
960x540& 31.65dB@0.089bpp& 33.74dB@0.071bpp\\
1920x1080& 31.56dB@0.023bpp& 33.21dB@0.018bpp\\
Enc. Time& 5h38min& 2h54min\\
\noalign{\hrule height 1.2pt}
\end{tabular}
\label{tab:scalable}
\end{table}
Scalable coding allows a single bitstream to be decoded at multiple quality levels or spatial resolutions. It is widely adopted in practical video delivery system.
However, existing INR-based methods are typically trained as resolution-specific models, limiting their applicability to scalable scenarios.
By introducing multi-resolution supervision, FaNeRV supports scalable decoding across different resolutions within a single model.
As shown in Tab. \ref{tab:scalable}, FaNeRV effectively captures multi-scale video information using one unified representation. In contrast, HiNeRV requires training separate models for each resolution and still incurs higher bitrates inferior reconstruction quality and longer training time.


\subsection{Ablation Studies}
To systematically assess the contribution of each component in our framework, we perform ablation studies on the HEVC ClassB \cite{sullivan2012overview} dataset.
As shown in Tab. ~\ref{tab:aba}, removing the Multi-Resolution Supervision (V1) leads to an 11.9\% BD-Rate increase, indicating that hierarchical frequency decomposition plays a critical role in balancing the optimization of global geometry and local textures.
Second, disabling the High-Frequency Injection (V2) results in a 7.5\% BD-Rate increase, demonstrating that explicitly injecting high-frequency residuals is crucial for mitigating spectral bias and preserving fine-grained textures.
Finally, we examine the upsampling mechanism by replacing our hybrid design with single operator variants. 
Using PixelShuffle exclusively (V3) causes a 13.9\% BD-rate increase, while relying solely on bilinear interpolation (V4) incurs a BD-rate penalty of 3.4\%. 
Our proposed hybrid strategy achieves an optimal balance between reconstruction quality and coding efficiency, attributed to its unique combination of trainable approaches and prior-based upsampling.
\label{subsec:ablation}

\begin{table}[t]
\centering
\caption{
Abalation Study on HEVC ClassB. The positive values of BD-Rate reflect the bitrate increases compared to FaNeRV.
}
\renewcommand{\arraystretch}{1.15}

\begin{tabular}{l|cccccc}
\noalign{\hrule height 1.2pt}
Methods & FaNeRV (Ours) & V1 & V2 & V3 & V4 \\
\noalign{\hrule height 0.8pt}
BD-rate (\%) & 0.00 & 11.9 & 7.5 & 13.9 & 3.4 \\
\noalign{\hrule height 1.2pt}
\end{tabular}

\label{tab:aba}
\end{table}
\section{Conclusion}
In this paper, we proposed FaNeRV, a frequency-aware implicit neural representation framework for video compression. FaNeRV reformulates video reconstruction as a progressive frequency-decoupled learning process, through multi-resolution supervision and dynamic high-frequency injection mechanism. In addition, we designed a network that decomposes intermediate features across different frequency bands, further enhancing the video representation capacity.
Extensive experiments on standard benchmarks demonstrated that FaNeRV outperforms existing INR-based and learning-based video compression methods in rate–distortion performance, and even surpasses the state-of-the-art hybrid codec VTM. In addition, FaNeRV supports scalable video coding within a single unified model, highlighting its potential for practical deployment.

\end{document}